\pdfoutput=1
\documentclass[11pt]{article}

\usepackage{emnlp2021}

\usepackage{times}
\usepackage{latexsym}
\usepackage{amsmath}
\usepackage{url}
\usepackage{amssymb}
\usepackage{amsfonts}
\usepackage{graphicx}
\usepackage{tabularx}
\usepackage{multirow}
\usepackage{arydshln}
\usepackage{mathtools,nccmath}
\usepackage{enumitem}
\usepackage{todonotes}
\usepackage{microtype}

\title{PhoMT: A High-Quality and Large-Scale Benchmark Dataset \\ for Vietnamese-English Machine Translation}

\author{Long Doan$^\ast$, Linh The Nguyen$^\ast$, Nguyen Luong Tran\thanks{\ \ The first three authors contributed equally to this work.} , \\ {\bf Thai Hoang}\thanks{\ \ Work done during internship at  VinAI Research. Email:  \texttt{qthai912@cs.washington.edu}} \and {\bf Dat Quoc Nguyen} \\
VinAI Research, Hanoi, Vietnam\\
\normalsize{\texttt{\{v.longdct, v.linhnt140, v.nguyentl12, v.thaihq5, v.datnq9\}@vinai.io}}
}

\begin{document}
\maketitle
\begin{abstract}
We introduce a high-quality and large-scale Vietnamese-English parallel dataset of 3.02M sentence pairs, which is 2.9M pairs larger than the benchmark Vietnamese-English machine translation corpus IWSLT15. We conduct experiments comparing strong neural baselines and well-known automatic translation engines on our dataset and find that in both automatic and human evaluations: the best performance is obtained by fine-tuning the pre-trained sequence-to-sequence denoising auto-encoder mBART. To our best knowledge, this is the first large-scale Vietnamese-English machine translation study. We hope our publicly available dataset and study can serve as a starting point for future research and applications on Vietnamese-English machine translation. We release our dataset at:  \url{https://github.com/VinAIResearch/PhoMT}.
\end{abstract}

\section{Introduction}

Vietnam has achieved rapid economic growth in the last two decades 
\cite{vietnamworkingpaper}. It is now an attractive destination for trade and investment. Due to the language barrier, foreigners usually rely on automatic machine translation (MT) systems to translate Vietnamese texts into their native language or another language they are familiar with, e.g. the global language English, so they could quickly catch up with ongoing events in Vietnam.  Thus the demand for high-quality  Vietnamese-English MT has rapidly increased. However, state-of-the-art MT models require high-quality and large-scale corpora for training to be able to reach near human-level translation quality \cite{45610,ott-etal-2018-scaling}.  
Despite being one of the most spoken languages in the world with about 100M speakers, Vietnamese is referred to as a low-resource language in MT research because publicly available parallel corpora for Vietnamese in general and in particular for Vietnamese-English MT are not large enough or have low-quality translation pairs,  including those with different sentence meaning (i.e. misalignment). 

Two main concerns are detailed as follows:

\begin{itemize}[leftmargin=*]
\item \textit{High-quality Vietnamese-English parallel corpora are either not publicly available or small-scale}. \citet{evbcorpus} and \citet{NMT_EN-VI_study} present two corpora each comprising of 800K sentence pairs, however, these two corpora are not publicly available. Thus, the Vietnamese-English parallel corpus IWSLT15 \citep{iwslt2015} of 133K sentence pairs extracted from TED-Talks transcripts is  still considered as the standard benchmark for MT when it comes to Vietnamese. Recently, the OPUS project \cite{tiedemann-2012-parallel} provides 300K+ sentence pairs extracted from the TED2020 v1 corpus of TED-Talks transcripts \cite{dataset_ted}. 

\item \textit{Larger Vietnamese-English parallel corpora are noisy},   
e.g. see discussions on the 300K-600K sentence pair corpora of JW300 \citep{jw300}, OPUS's GNOME and QED \citep{dataset_qed} in Section \ref{ssec:parallel}, and on the OpenSubtitles corpus \citep{dataset_opensub} in Section \ref{ssec:pre}. 
Recently, 
CCAligned \citep{ccaligned} and WikiMatrix \cite{wikimatrix} are created by using LASER sentence embeddings  \cite{laser} and margin-based sentence alignment to mine parallel sentences from comparable web-document pairs.
Though containing millions of Vietnamese-English parallel sentence pairs, they still have a large proportion of misalignment and low-quality translation pairs. In particular, we randomly sample from each corpus 100 sentence pairs and manually inspect their quality. We find that only 37/100 CCAligned pairs and 31/100 WikiMatrix pairs are at a high-quality translation level. 
\end{itemize}

As the \textit{first contribution}, to help handle the two concerns above, we present a high-quality and large-scale Vietnamese-English parallel dataset, named \textbf{PhoMT}, that consists of 3.02M sentence pairs.  Here, from PhoMT, we also prepare 38K sentence pairs with manually qualitative inspection, that are used for validation and test. We believe that our dataset construction process will help develop more efficient data creation strategies for other low-resource languages. %\todo{Add state that our approach to build the corpus can be applied for other languages.}
As the \textit{second contribution}, we empirically investigate strong baselines on our dataset, including Transformer-base, Transformer-big \citep{transformer}  and the pre-trained sequence-to-sequence denoising auto-encoder mBART  \citep{mbart}, and compare these baselines with well-known automatic translation engines.
We find that mBART obtains the highest scores in terms of both automatic and human evaluations on both translation directions. To the best of our knowledge, this is the first large-scale empirical study for Vietnamese-English MT. As our \textit{final contribution},  we publicly release our PhoMT dataset for research or educational purposes.  We hope PhoMT together with  our empirical study can serve as a starting point for future Vietnamese-English MT research and applications.

\section{Our PhoMT dataset}

Our dataset construction process consists of 4 phases. The 1st phase is to collect parallel document pairs. The 2nd phase is a pre-processing  step that is to produce cleaned and high-quality parallel document pairs and then extract  sentences from these pairs. The 3rd phase is to align parallel sentences within a pair of parallel documents. The 4th phase is a post-processing step that is to filter out duplicated parallel sentence pairs and manually verify the quality of validation and test sets.   

\subsection{Collecting parallel document pairs}
\label{ssec:parallel}

We collect the parallel document pairs from publicly available resources that contain original English documents and their corresponding Vietnamese-translated version. \underline{\textbf{WikiHow}}: It is an online knowledge base of how-to guides that are available in multiple languages. We employ a multilingual WikiHow-based document summarization corpus \cite{dataset_wikihow} that contains 6616 pairs of WikiHow English articles and their Vietnamese-translated variant.   \underline{\textbf{TED-Talks}}: We use the TED2020 v1 corpus \cite{dataset_ted} that includes 3123 English-Vietnamese subtitle pairs of TED talks. 
\underline{\textbf{OpenSubtitles}}: We employ the latest version v2018 of the OpenSubtitles corpus \citep{dataset_opensub} that contains 3886 parallel movie and TV subtitles. %Here, we download both the TED2020 and OpenSubtitles corpora from OPUS \cite{tiedemann-2012-parallel}. 
\underline{\textbf{MediaWiki}}: We also use parallel documents from the MediaWiki content translation data dump. \underline{\textbf{News}} \& \underline{\textbf{Blogspot}}:  We collect  English and Vietnamese-translated versions of news and Blogspot articles  from eight websites for English learners. See URLs for the described resources in the Appendix. 

Here, we do not include available corpora of JW300  \citep{jw300}, OPUS's GNOME and QED \citep{dataset_qed}. We manually check 100 randomly sampled pairs from the 600K Vietnamese-English sentence pair corpus JW300 and find that there are 71 high-quality translation pairs. However, it is worth noting that JW300 can introduce potential bias because of its religious domain. GNOME from OPUS contains 600K sentences pairs, in which most Vietnamese target sentences include many original translatable technical English words, thus not natural. QED  has 340K sentence pairs, however, our investigation finds that about a half of the QED pairs are from the TED-Talks transcripts \cite{dataset_ted}; and from the remaining sentence pairs, we randomly sample 100 pairs and find that only 43 pairs have a high-quality translation.

\begin{table*}[!t]
    \centering
    \resizebox{15.5cm}{!}{
    \begin{tabular}{l|l|l|l|l|l|l|l|l|l|l|l}
    \hline
        \multirow{2}{*}{\textbf{Domain}} & \multicolumn{2}{c|}{\textbf{Total}} & \multicolumn{3}{c|}{\textbf{Training}} & \multicolumn{3}{c|}{\textbf{Validation}} & \multicolumn{3}{c}{\textbf{Test}} \\
        \cline{2-12}
         & \textbf{\#doc} & \textbf{\#pair}
         & \textbf{\#pair} & \textbf{\#en/s} & \textbf{\#vi/s} 
         & \textbf{\#pair} & \textbf{\#en/s} & \textbf{\#vi/s}
         & \textbf{\#pair} & \textbf{\#en/s} & \textbf{\#vi/s}
         \\
        \hline
        News & 2559 & 41504 & 40990 & 24.4 & 32.0 & 257 & 22.3 & 30.3 & 257 & 26.8 & 34.5 \\
        Blogspot & 1071 & 93956 & 92545 & 25.0 & 34.6 & 597 & 26.4 & 37.8 & 814 & 23.7 & 31.5 \\
        TED-Talks & 3123 & 320802 & 316808 & 19.8 & 23.8 & 1994 & 20.0 & 24.6 & 2000 & 22.0 & 27.9 \\
        MediaWiki & 38969 & 496799 & 490505 & 26.0 & 32.8 & 3024 & 25.3 & 32.3 & 3270 & 27.0 & 33.7\\ 
        WikiHow & 6616 & 513837 & 507379 & 18.9 & 22.4 & 3212 & 17.9 & 21.5 & 3246 & 17.5 & 21.5 \\
        OpenSub & 3312 & 1548971 & 1529772 & 9.7 & 11.1 & 9635 & 9.5 & 10.7 & 9564 & 10.0 & 11.4 \\
        \hline
        All & 55650 & 3015869 & 2977999 & 15.7 & 19.0 & 18719 & 15.3 & 18.7 & 19151 & 16.2 & 19.8 \\
    \hline
    \end{tabular}
    }
    \caption{Our dataset statistics. ``\#doc'',  ``\#pair'', ``\#en/s'' and ``\#vi/s'' denote the number of parallel document pairs, the number of aligned parallel sentence pairs,  the average number of word tokens per English sentence and the average number of syllable tokens per Vietnamese sentence, respectively.
    ``OpenSub'' abbreviates OpenSubtitles.}
    \label{tab:corpus_statistics}
\end{table*}

\subsection{Pre-processing}\label{ssec:pre}

We find that not all of 3886 English-Vietnamese parallel document pairs in OpenSubtitles have a high-quality translation. We manually inspect each OpenSubtitles pair and remove 574/3886 (15\%) document pairs with a low-quality translation, thus remaining 3312  pairs. 
In MediaWiki, there are original English paragraphs appearing in some Vietnamese target documents, that have not been translated into Vietnamese yet. We employ the language identification module  of fastText \citep{joulin-etal-2017-bag} to identify and filter those English paragraphs out of the Vietnamese documents. We also remove reference sections and tables that appear in some MediaWiki and Blogspot documents. 

To extract sentences for parallel sentence alignment in the next phase, we perform (tokenization and) sentence segmentation by using the Stanford CoreNLP toolkit \citep{corenlp} and RDRSegmenter \cite{nguyen-etal-2018-fast} from the VnCoreNLP toolkit \citep{vncorenlp} for English and Vietnamese, respectively. %\todo{cite RDRSegmenter}

\subsection{Aligning parallel sentence pairs}

To align parallel sentences within a parallel document pair, our approach first employs the strong neural MT engine Google Translate to translate each English source sentence into Vietnamese. %The reason why we use this English-to-Vietnamese translation direction is because Google-translating from English to Vietnamese produces better translation quality than from Vietnamese to English \cite{kudo-2018-subword}. 
Then we use three toolkits of Hunalign \citep{hunalign}, Gargantua \citep{gargantua} and Bleualign \citep{bleualign} to perform sentence alignment via alignment between the Google-translated variants of the English source sentences and the Vietnamese target sentences. Finally, we only select sentence pairs that are aligned by at least two out of three toolkits as the output of our alignment process.

The quality of our sentence alignment output is shown in Section \ref{ssec:post}. Here, we discuss alignment coverage rates.  On the same TED2020 v1 corpus, the  automatic alignment approach OPUS \cite{tiedemann-2012-parallel}, based on word alignments and phrase tables, aligns a total of 326K Vietnamese sentences,\footnote{\url{https://object.pouta.csc.fi/OPUS-TED2020/v1/moses/en-vi.txt.zip} wherein duplicate removal is not performed.} 
while our approach aligns 350K Vietnamese ones (i.e. a 7.5\% relative improvement).\footnote{See the Appendix for an additional discussion.} 

Note that from each resource domain except OpenSubtitles, our approach selects 99+\% of Vietnamese sentences to be included in the output of our alignment process. Particularly, a total of 14.6K Vietnamese sentences (i.e. 0.86\%) are not selected from five resource domains of News, Blogspot, TED-Talks, MediaWiki and WikiHow. When it comes to OpenSubtitles, the rate reduces to 95\% (here, 120K Vietnamese sentences are not included in our alignment output). 
On average, the alignment coverage rate for English is about 2\% absolute lower than the one for Vietnamese as there are English source sentences that are not translated to Vietnamese in the collected corpora.

\begin{table*}[!t]
\centering
\resizebox{15.5cm}{!}{
\begin{tabular}{l|c|c|c|c|c|c|c|c|c|c}
\hline
\multirow{3}{*}{\textbf{Model}} & \multicolumn{4}{c|}{\textbf{Validation set}}& \multicolumn{6}{c}{\textbf{Test set}} \\
\cline{2-11}
 & \multicolumn{2}{c|}{\textbf{En-to-Vi}}& \multicolumn{2}{c|}{\textbf{Vi-to-En}}& \multicolumn{3}{c|}{\textbf{En-to-Vi}}& \multicolumn{3}{c}{\textbf{Vi-to-En}} \\
\cline{2-11}
& \textbf{TER}$\downarrow$ & \textbf{BLEU}$\uparrow$   
& \textbf{TER}$\downarrow$ & \textbf{BLEU}$\uparrow$   
& \textbf{TER}$\downarrow$ & \textbf{BLEU}$\uparrow$  & \textbf{Human}$\uparrow$ 
& \textbf{TER}$\downarrow$ & \textbf{BLEU}$\uparrow$  & \textbf{Human}$\uparrow$ 
\\
\hline
Google Translate & 45.86 & 40.10 & 44.69 & 36.89 & 46.52 & 39.86 & 23/100  & 45.86 & 35.76 & 10/100  \\
Bing Translator & 45.36 & 40.82 & 45.32 & 36.61 & 46.04 & 40.37 & 14/100  & 46.09 & 35.74 &  15/100 \\
\hdashline
Transformer-base & 42.77 & 43.01 & 43.42 & 38.26 & 43.79 & 42.12 & 13/100  & 44.28 & 37.19 &  13/100  \\
Transformer-big & 42.13 & 43.75 & 43.08 & 39.04 & 43.04 & 42.94 &  18/100 & 44.06 & 37.83 & 28/100  \\
mBART & \textbf{41.56} & \textbf{44.32} & \textbf{41.44} & \textbf{40.88} & \textbf{42.57} & \textbf{43.46} &  \textbf{32/100}  & \textbf{42.54} & \textbf{39.78} & \textbf{34/100}  \\
\hline
\end{tabular}
}
\caption{Overall results. Each TER/BLEU score difference between two models is statistically significant  (p-value $<$ 0.05 based on bootstrap resampling), except Google Translate and Bing Translator for Vi-to-En w.r.t. BLEU. }
\label{tab:results}
\end{table*}

\subsection{Post-processing}\label{ssec:post}

On  the alignment output from the previous phase, we normalize punctuations, remove all duplicate sentence pairs within and across all document pairs, and also remove sentence pairs where the reference English sentence contains either only a single word token or swear words.\footnote{Regarding removals, punctuations are not taken into identifying sentence pair duplication or computing sentence length. On OpenSubtitles, we also remove sentence pairs where the reference English sentence consists of two tokens.}   
Then we randomly split each domain into training/validation/test sets on document level with a  98.75/0.60/0.65 ratio, resulting in 2977999 training, 18876 validation and 19291 test sentence pairs.

To qualify our dataset, we  manually inspect each sentence pair in the validation and test sets. Here, each pair is inspected by two out of the first three co-authors independently: one inspector checks about (18876 + 19291) $\times$ 2 / 3 = 25K sentence pairs in 125 hours on average (i.e. 200 sentences/hour). 
After cross-checking, we find that in the validation set, 32 sentence pairs (0.17\%) are misaligned (i.e. completely different sentence meaning or partly preserving the sentence meaning); and 125 pairs (0.66\%) are low-quality translation ones (i.e. mostly or completely preserving the sentence meaning, however, the Vietnamese target sentence is not naturally smooth). In the test set, there are 27 misaligned sentence pairs (0.14\%) and 113 low-quality translation pairs (0.59\%). 
Note that performing a similar manual inspection on the training set is beyond our current human resource; however, with small proportions of misalignment and low-quality translation on the validation and test sets at the sentence pair level, 
we believe that our training set attains a high-quality standard. Lastly, we remove those 32 + 125 + 27 + 113 = 297  pairs, resulting in 18719 validation and 19151 test sentence pairs of high quality for final use.

\section{Experiments}

\subsection{Experimental setup}

We conduct experiments on our PhoMT dataset to study: (i) a comparison between the well-known automatic translation engines (here, Google Translate and Bing Translator) and strong neural MT baselines, and (ii) the usefulness of pre-trained sequence-to-sequence denoising auto-encoder. In particular, we use the baseline models: Transformer-base, Transformer-big \citep{transformer}, %BERT-fused \citep{bert-nmt}, 
and the pre-trained denoising auto-encoder mBART  \citep{mbart}.  
%Following \citet{bert-nmt}, we initialize BERT-fused with our best Transformer-base checkpoint. Unlike \citet{bert-nmt}, our BERT-fused incorporates  the pre-trained multilingual language model XLM-R \citep{xlm-r}  instead of  XLM\textsubscript{MLM+TLM} \cite{xlmtlm} as XLM\textsubscript{MLM+TLM} is pre-trained using corpora that include all Vietnamese documents from the OpenSubtitles v2018 dataset. 
We report standard metrics  TER \citep{ter_metrics} and BLEU \citep{bleu_metrics}, in which lower TER and higher BLEU indicate better performances. We compute the case-sensitive BLEU score using SacreBLEU \cite{sacrebleu_metrics}. 
See the Appendix for implementation details. Here, we select the model checkpoint that obtains the highest BLEU score on the validation set to apply to the test set.

\subsection{Automatic evaluation results}

Table \ref{tab:results} presents TER and BLEU scores obtained by the automatic translation engines and the neural baselines on the validation and test sets for the English-to-Vietnamese (En-to-Vi) and Vietnamese-to-English (Vi-to-En) translation setups. Clearly, the baselines obtain significantly  better TER and BLEU scores than the automatic translation engines  for both En-to-Vi and Vi-to-En setups on both validation and test sets. It is likely because Google Translate and Bing Translator are trained on other parallel resources. Here, Transformer models obtain 
1.5+ points absolute better than both Google Translate and Bing Translator, in which Transformer-big outperforms Transformer-base. In addition, mBART achieves the best performance among all models, reconfirming the quantitative effectiveness of multilingual denoising pre-training for neural MT \citep{mbart}.

\begin{table}[!t]
\centering
\resizebox{7.5cm}{!}{
\setlength{\tabcolsep}{0.3em}
\begin{tabular}{l|c|c|c|c|c|c}
\hline
\multirow{2}{*}{\textbf{Model}} &   \multicolumn{6}{c}{\textbf{Vi-to-En}} \\
\cline{2-7}
 & {\textbf{News}} & {\textbf{BloS}} & {\textbf{TedT}} & {\textbf{MedW}} & {\textbf{WikH}} & {\textbf{OpeS}}  \\
\hline
Google Translate & 34.33 & 26.71 & 34.03 & 48.81 & 28.76 & 28.63\\
Bing Translator  & 35.05 & 25.91 & 32.03 & 50.29 & 24.78 & 30.56\\
 \hdashline
Transformer-base & 34.94 & 27.13 & 33.46 & 50.56 & 28.37 & 32.29\\
Transformer-big  & 36.14 & 28.26 & 34.46 & 51.20 & 28.86 & 32.69\\
    mBART        & \textbf{37.04} & \textbf{29.26} & \textbf{36.43} & \textbf{53.93} & \textbf{30.63} & \textbf{34.13}\\
\hline
\end{tabular}
}
\caption{BLEU scores for each resource domain on the test set. Here, ``BloS'', ``TedT'', ``MedW'', ``WikH'' and ``OpeS'' abbreviate Blogspot, TED-Talks, MediaWiki, WikiHow and OpenSubtitles, respectively.}
\label{tab:bleu_domain_test}
\end{table}

\begin{table}[!t]
\centering
\resizebox{7.5cm}{!}{
\setlength{\tabcolsep}{0.3em}
\begin{tabular}{l|c|c|c|c|c|c}
\hline
\multirow{3}{*}{\textbf{Model}} & \multicolumn{6}{c}{\textbf{Vi-to-En}} \\
\cline{2-7}
& {\textbf{$<$ 10}} & {\textbf{[10, 20)}} & {\textbf{[20, 30)}} & {\textbf{[30, 40)}} & {\textbf{[40, 50)}} & {\textbf{$\geq$ 50}}  \\
%\cline{2-7}
& 48.77\% & 30.87\% & 13.65\% & 4.46\% & 1.45\% & 0.80\% \\
\hline
 Google Translate & 32.27 & 34.24 & 37.47 & 38.17 & 37.94 & 38.30 \\
 Bing Translator &  34.20 & 32.97 & 36.66 & 38.27 & 39.73 & 40.70 \\
 \hdashline
 Transformer-base & 36.78 & 34.99 & 38.27 & 38.86 & 38.76 & 39.46  \\
 Transformer-big &  37.23 & 35.49 & 39.02 & 39.90 & 39.34 & 39.85 \\
 mBART & \textbf{38.31} & \textbf{37.22} & \textbf{41.05} & \textbf{42.21} & \textbf{42.49} & \textbf{43.35}  \\ 
\hline
\end{tabular}
}
\caption{BLEU scores on the test set w.r.t. sentence lengths of reference English sentences (i.e. the number of word tokens including punctuations). The number right below each length bucket denotes the percentage of sentences belonging to the bucket.}
\label{tab:bleu_sent_len_vien}
\end{table}

We present BLEU scores on the Vi-to-En test set for each resource domain and sentence length bucket in Tables \ref{tab:bleu_domain_test} and \ref{tab:bleu_sent_len_vien}, respectively. 
Table \ref{tab:bleu_domain_test} shows that the highest BLEU scores are reported for MediaWiki (wherein documents share and link common events or topics), followed by the ones reported for News and TED-Talks. Three remaining resource domains Blogspot, WikiHow and OpenSubtitles contain less common topic-specific document pairs, thus resulting in lower scores. 
In Table \ref{tab:bleu_sent_len_vien}, we find that models produce lower BLEU scores for short- and medium-length sentences (i.e. $<$ 20 tokens) than for long sentences. This is not surprising as a major proportion of short- and medium-length sentences are from OpenSubtitles, while longer sentences generally come from MediaWiki, News and TED-Talks. Note that we observe similar findings for the TER results as well as on the validation set and in the En-to-Vi setup. See additional results w.r.t. each domain and sentence length in the Appendix.

\begin{figure}[!t]
	\centering
	\includegraphics[width=0.9\linewidth]{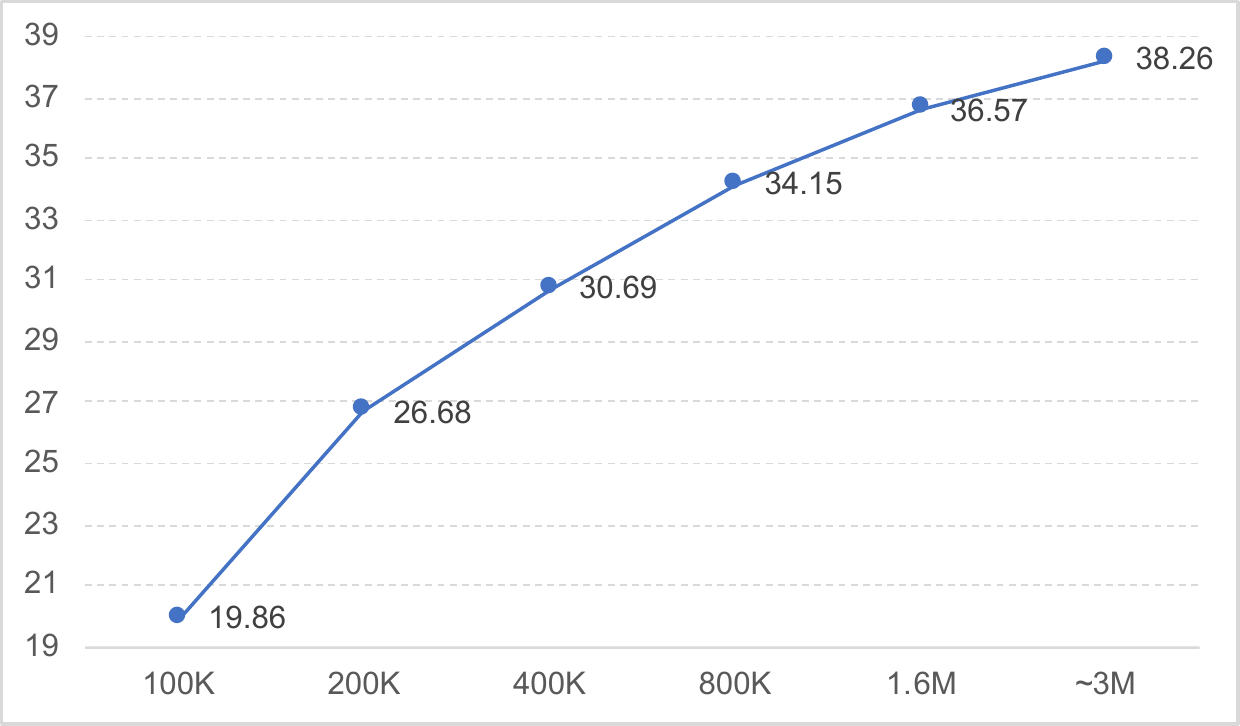}
    \caption{BLEU scores of Transformer-base on the Vi-to-En validation set when varying training sizes.}
    \label{fig:bleu_trainingsize}
 \end{figure}

Figure \ref{fig:bleu_trainingsize} presents BLEU scores of Transformer-base on the validation set  for the Vi-to-En setup when varying the numbers of training sentence pairs. Those scores clearly show the effectiveness of larger training sizes.

We also perform an experiment to additionally show that our curation effort has paid off. 
In particular, as not all of our data are overlapping with OPUS, for a fair comparison, we sample a set of 1.55M non-duplicate Vietnamese-English sentence pairs from OPUS’s OpenSubtitles, which has the same size as our PhoMT's OpenSubtitles training subset and do not contain pairs appearing in our OpenSubtitles validation and test subsets. 
We train two Transformer-base models for Vi-to-En translation: one trained using
the sampled OPUS’s OpenSubtitles set and another one trained using our OpenSubtitles training subset. Hyper-parameter tuning is performed using our OpenSubtitles validation subset in the same manner as presented in the Appendix. We evaluate the models using our OpenSubtitles test subset. 
We find that Transformer-base trained using the sampled OPUS’s OpenSubtitles set produces a significantly lower Vi-to-En BLEU score on our OpenSubtitles test subset than the one trained using our OpenSubtitles training subset (29.72 vs. 31.11), clearly showing the effectiveness of our quality control. Note that as shown in Table \ref{tab:bleu_domain_test}, Transformer-base trained using the whole PhoMT's training set obtains a higher Vi-to-En BLEU score at 32.29 on our OpenSubtitles test subset. Thus this experiment  also reconfirms the positive effect of a larger training size.

\subsection{Human evaluation results}

We also conduct a human-based manual comparison between the outputs generated by the two automatic translation systems and our three neural baselines. For each translation direction, we randomly sample 100 source sentences in the test set; and for each sentence sample, we anonymously shuffle the translation outputs  from five systems. Here, each sampled pair satisfies that any two out of five translation outputs are not exactly the same. We then ask three external Vietnamese annotators to choose which translation  they think is the best.\footnote{Annotators have a proficient English level at IELTS 7.0+ with a reading score of 8.0+ (paid 0.25USD per sample).} 
The inter-annotator agreement score computed for Fleiss' kappa coefficient \cite{fleiss1971measuring} between the three annotators is 0.63 which is relatively substantial. Our fourth co-author hosts and participates in a discussion session  with the three annotators to resolve annotation conflicts.\footnote{The fourth co-author is not involved in both the post-processing step (Section \ref{ssec:post}) and the development of MT systems. He does not know which model produces which translation. He also has a proficient English level.} 

Table \ref{tab:results} shows final results, where mBART gains the highest  human evaluation scores, thus demonstrating its qualitative effectiveness  for both En-to-Vi and Vi-to-En translation. Table \ref{tab:results}  also shows that human preference is not always correlated with the automatic  metrics TER and BLEU. For example, in the En-to-Vi setup, though Transformer models have 
2+ points better TER and BLEU than Google Translate, they are less preferred by humans than Google Translate (13 vs. 23 and 18 vs. 23). A detailed study is beyond the scope of our paper, but it is worth investigating in future work.

\section{Conclusion}

We have presented PhoMT---a high-quality and large-scale Vietnamese-English parallel dataset of 3.02M sentence pairs. We empirically conduct experiments on our PhoMT dataset to compare strong baselines and demonstrate the effectiveness of the pre-trained denoising auto-encoder mBART for neural MT in both automatic and human evaluations. We hope that the public release of our dataset 
can serve as the starting point for further Vietnamese-English MT research and applications. %Future work involves filtering low-quality translation sentence pairs from existing parallel corpora. 

\section*{Acknowledgments}
The authors would like to thank the anonymous reviewers for  their helpful feedback.

% Entries for the entire Anthology, followed by custom entries
\bibliography{refs}

\begin{thebibliography}{32}
\expandafter\ifx\csname natexlab\endcsname\relax\def\natexlab#1{#1}\fi

\bibitem[{Abdelali et~al.(2014)Abdelali, Guzman, Sajjad, and
  Vogel}]{dataset_qed}
Ahmed Abdelali, Francisco Guzman, Hassan Sajjad, and Stephan Vogel. 2014.
\newblock {The {AMARA} Corpus: Building Parallel Language Resources for the
  Educational Domain}.
\newblock In \emph{Proceedings of the Ninth International Conference on
  Language Resources and Evaluation}, pages 1856--1862.

\bibitem[{Agi{\'c} and Vuli{\'c}(2019)}]{jw300}
{\v{Z}}eljko Agi{\'c} and Ivan Vuli{\'c}. 2019.
\newblock {{JW}300: A Wide-Coverage Parallel Corpus for Low-Resource
  Languages}.
\newblock In \emph{Proceedings of the 57th Annual Meeting of the Association
  for Computational Linguistics}, pages 3204--3210.

\bibitem[{Artetxe and Schwenk(2019)}]{laser}
Mikel Artetxe and Holger Schwenk. 2019.
\newblock Massively multilingual sentence embeddings for zero-shot
  cross-lingual transfer and beyond.
\newblock \emph{Transactions of the Association for Computational Linguistics},
  7:597--610.

\bibitem[{Baum(2020)}]{vietnamworkingpaper}
Anja Baum. 2020.
\newblock {Vietnam's Development Success Story and the Unfinished SDG Agenda}.
\newblock \emph{IMF Working Papers}, 20(31):1--31.

\bibitem[{Braune and Fraser(2010)}]{gargantua}
Fabienne Braune and Alexander Fraser. 2010.
\newblock {Improved Unsupervised Sentence Alignment for Symmetrical and
  Asymmetrical Parallel Corpora}.
\newblock In \emph{Proceedings of the 23rd International Conference on
  Computational Linguistics: Posters}, pages 81--89.

\bibitem[{Cettolo et~al.(2015)Cettolo, Niehues, St{\"u}ker, Bentivogli,
  Cattoni, and Federico}]{iwslt2015}
M.~Cettolo, J.~Niehues, S.~St{\"u}ker, L.~Bentivogli, R.~Cattoni, and Marcello
  Federico. 2015.
\newblock {The IWSLT 2015 Evaluation Campaign}.
\newblock In \emph{Proceedings of the International Workshop on Spoken Language
  Translation}.

\bibitem[{Ding et~al.(2019)Ding, Renduchintala, and Duh}]{subword_size}
Shuoyang Ding, Adithya Renduchintala, and Kevin Duh. 2019.
\newblock {A Call for Prudent Choice of Subword Merge Operations in Neural
  Machine Translation}.
\newblock In \emph{Proceedings of Machine Translation Summit XVII Volume 1:
  Research Track}, pages 204--213.

\bibitem[{El-Kishky et~al.(2020)El-Kishky, Chaudhary, Guzm{\'a}n, and
  Koehn}]{ccaligned}
Ahmed El-Kishky, Vishrav Chaudhary, Francisco Guzm{\'a}n, and Philipp Koehn.
  2020.
\newblock {{CCA}ligned: A Massive Collection of Cross-Lingual Web-Document
  Pairs}.
\newblock In \emph{Proceedings of the 2020 Conference on Empirical Methods in
  Natural Language Processing}, pages 5960--5969.

\bibitem[{Fleiss(1971)}]{fleiss1971measuring}
Joseph~L Fleiss. 1971.
\newblock Measuring nominal scale agreement among many raters.
\newblock \emph{Psychological bulletin}, 76(5):378.

\bibitem[{Joulin et~al.(2017)Joulin, Grave, Bojanowski, and
  Mikolov}]{joulin-etal-2017-bag}
Armand Joulin, Edouard Grave, Piotr Bojanowski, and Tomas Mikolov. 2017.
\newblock {Bag of Tricks for Efficient Text Classification}.
\newblock In \emph{Proceedings of the 15th Conference of the {E}uropean Chapter
  of the Association for Computational Linguistics: Volume 2, Short Papers},
  pages 427--431.

\bibitem[{Kingma and Ba(2014)}]{KingmaB14}
Diederik~P. Kingma and Jimmy Ba. 2014.
\newblock {Adam: {A} Method for Stochastic Optimization}.
\newblock \emph{arXiv preprint}, arXiv:1412.6980.

\bibitem[{Ladhak et~al.(2020)Ladhak, Durmus, Cardie, and
  McKeown}]{dataset_wikihow}
Faisal Ladhak, Esin Durmus, Claire Cardie, and Kathleen McKeown. 2020.
\newblock {{W}iki{L}ingua: A New Benchmark Dataset for Cross-Lingual
  Abstractive Summarization}.
\newblock In \emph{Findings of the Association for Computational Linguistics:
  EMNLP 2020}, pages 4034--4048.

\bibitem[{Lison and Tiedemann(2016)}]{dataset_opensub}
Pierre Lison and J{\"o}rg Tiedemann. 2016.
\newblock {{O}pen{S}ubtitles2016: Extracting Large Parallel Corpora from Movie
  and {TV} Subtitles}.
\newblock In \emph{Proceedings of the Tenth International Conference on
  Language Resources and Evaluation}, pages 923--929.

\bibitem[{Liu et~al.(2020)Liu, Gu, Goyal, Li, Edunov, Ghazvininejad, Lewis, and
  Zettlemoyer}]{mbart}
Yinhan Liu, Jiatao Gu, Naman Goyal, Xian Li, Sergey Edunov, Marjan
  Ghazvininejad, Mike Lewis, and Luke Zettlemoyer. 2020.
\newblock {Multilingual Denoising Pre-training for Neural Machine Translation}.
\newblock \emph{Transactions of the Association for Computational Linguistics},
  8:726--742.

\bibitem[{Manning et~al.(2014)Manning, Surdeanu, Bauer, Finkel, Bethard, and
  McClosky}]{corenlp}
Christopher Manning, Mihai Surdeanu, John Bauer, Jenny Finkel, Steven Bethard,
  and David McClosky. 2014.
\newblock {The {S}tanford {C}ore{NLP} Natural Language Processing Toolkit}.
\newblock In \emph{Proceedings of 52nd Annual Meeting of the Association for
  Computational Linguistics: System Demonstrations}, pages 55--60.

\bibitem[{Ngo et~al.(2013)Ngo, Winiwarter, and Wloka}]{evbcorpus}
Quoc~Hung Ngo, Werner Winiwarter, and Bartholom{\"a}us Wloka. 2013.
\newblock {{EVBC}orpus - A Multi-Layer {E}nglish-{V}ietnamese Bilingual Corpus
  for Studying Tasks in Comparative Linguistics}.
\newblock In \emph{Proceedings of the 11th Workshop on {A}sian Language
  Resources}, pages 1--9.

\bibitem[{Nguyen et~al.(2018)Nguyen, Nguyen, Vu, Dras, and
  Johnson}]{nguyen-etal-2018-fast}
Dat~Quoc Nguyen, Dai~Quoc Nguyen, Thanh Vu, Mark Dras, and Mark Johnson. 2018.
\newblock {A Fast and Accurate Vietnamese Word Segmenter}.
\newblock In \emph{Proceedings of the 11th International Conference on Language
  Resources and Evaluation}, pages 2582--2587.

\bibitem[{Ott et~al.(2019)Ott, Edunov, Baevski, Fan, Gross, Ng, Grangier, and
  Auli}]{fairseq}
Myle Ott, Sergey Edunov, Alexei Baevski, Angela Fan, Sam Gross, Nathan Ng,
  David Grangier, and Michael Auli. 2019.
\newblock {fairseq: A Fast, Extensible Toolkit for Sequence Modeling}.
\newblock In \emph{Proceedings of the 2019 Conference of the North American
  Chapter of the Association for Computational Linguistics: Demonstrations},
  pages 48--53.

\bibitem[{Ott et~al.(2018)Ott, Edunov, Grangier, and
  Auli}]{ott-etal-2018-scaling}
Myle Ott, Sergey Edunov, David Grangier, and Michael Auli. 2018.
\newblock {Scaling Neural Machine Translation}.
\newblock In \emph{Proceedings of the Third Conference on Machine Translation:
  Research Papers}, pages 1--9.

\bibitem[{Papineni et~al.(2002)Papineni, Roukos, Ward, and Zhu}]{bleu_metrics}
Kishore Papineni, Salim Roukos, Todd Ward, and Wei-Jing Zhu. 2002.
\newblock {{B}leu: a Method for Automatic Evaluation of Machine Translation}.
\newblock In \emph{Proceedings of the 40th Annual Meeting of the Association
  for Computational Linguistics}, pages 311--318.

\bibitem[{Phan-Vu et~al.(2019)Phan-Vu, Tran, Nguyen, Dang, and
  Do}]{NMT_EN-VI_study}
Hong-Hai Phan-Vu, Viet~Trung Tran, Van~Nam Nguyen, Hoang~Vu Dang, and
  Phan~Thuan Do. 2019.
\newblock {Neural Machine Translation between Vietnamese and English: an
  Empirical Study}.
\newblock \emph{Journal of Computer Science and Cybernetics}, 35(2):147--166.

\bibitem[{Post(2018)}]{sacrebleu_metrics}
Matt Post. 2018.
\newblock {A Call for Clarity in Reporting {BLEU} Scores}.
\newblock In \emph{Proceedings of the Third Conference on Machine Translation:
  Research Papers}, pages 186--191.

\bibitem[{Reimers and Gurevych(2020)}]{dataset_ted}
Nils Reimers and Iryna Gurevych. 2020.
\newblock {Making Monolingual Sentence Embeddings Multilingual using Knowledge
  Distillation}.
\newblock In \emph{Proceedings of the 2020 Conference on Empirical Methods in
  Natural Language Processing}, pages 4512--4525.

\bibitem[{Schwenk et~al.(2021)Schwenk, Chaudhary, Sun, Gong, and
  Guzm{\'a}n}]{wikimatrix}
Holger Schwenk, Vishrav Chaudhary, Shuo Sun, Hongyu Gong, and Francisco
  Guzm{\'a}n. 2021.
\newblock {{W}iki{M}atrix: Mining 135{M} Parallel Sentences in 1620 Language
  Pairs from {W}ikipedia}.
\newblock In \emph{Proceedings of the 16th Conference of the European Chapter
  of the Association for Computational Linguistics: Main Volume}, pages
  1351--1361.

\bibitem[{Sennrich et~al.(2016)Sennrich, Haddow, and Birch}]{subword_bpe}
Rico Sennrich, Barry Haddow, and Alexandra Birch. 2016.
\newblock {Neural Machine Translation of Rare Words with Subword Units}.
\newblock In \emph{Proceedings of the 54th Annual Meeting of the Association
  for Computational Linguistics (Volume 1: Long Papers)}, pages 1715--1725.

\bibitem[{Sennrich and Volk(2011)}]{bleualign}
Rico Sennrich and Martin Volk. 2011.
\newblock {Iterative, {MT}-based Sentence Alignment of Parallel Texts}.
\newblock In \emph{Proceedings of the 18th Nordic Conference of Computational
  Linguistics}, pages 175--182.

\bibitem[{Snover et~al.(2006)Snover, Dorr, Schwartz, Micciulla, and
  Makhoul}]{ter_metrics}
Matthew Snover, Bonnie Dorr, Richard Schwartz, Linnea Micciulla, and John
  Makhoul. 2006.
\newblock A study of translation edit rate with targeted human annotation.
\newblock In \emph{Proceedings of the 7th Biennial Conference of the
  Association for Machine Translation in the Americas}, pages 223--231.

\bibitem[{Tiedemann(2012)}]{tiedemann-2012-parallel}
J{\"o}rg Tiedemann. 2012.
\newblock {Parallel Data, Tools and Interfaces in {OPUS}}.
\newblock In \emph{Proceedings of the Eighth International Conference on
  Language Resources and Evaluation}, pages 2214--2218.

\bibitem[{Varga et~al.(2007)Varga, Hal{\'a}csy, Kornai, Nagy, N{\'e}meth, and
  Tr{\'o}n}]{hunalign}
D{\'a}niel Varga, P{\'e}ter Hal{\'a}csy, Andr{\'a}s Kornai, Viktor Nagy,
  L{\'a}szl{\'o} N{\'e}meth, and Viktor Tr{\'o}n. 2007.
\newblock {Parallel corpora for medium density languages}.
\newblock \emph{Current Issues in Linguistic Theory}, 292:247--258.

\bibitem[{Vaswani et~al.(2017)Vaswani, Shazeer, Parmar, Uszkoreit, Jones,
  Gomez, Kaiser, and Polosukhin}]{transformer}
Ashish Vaswani, Noam Shazeer, Niki Parmar, Jakob Uszkoreit, Llion Jones,
  Aidan~N Gomez, \L~ukasz Kaiser, and Illia Polosukhin. 2017.
\newblock {Attention is All you Need}.
\newblock In \emph{Advances in Neural Information Processing Systems},
  volume~30, pages 5998--6008.

\bibitem[{Vu et~al.(2018)Vu, Nguyen, Nguyen, Dras, and Johnson}]{vncorenlp}
Thanh Vu, Dat~Quoc Nguyen, Dai~Quoc Nguyen, Mark Dras, and Mark Johnson. 2018.
\newblock {{V}n{C}ore{NLP}: A {V}ietnamese Natural Language Processing
  Toolkit}.
\newblock In \emph{Proceedings of the 2018 Conference of the North {A}merican
  Chapter of the Association for Computational Linguistics: Demonstrations},
  pages 56--60.

\bibitem[{Wu et~al.(2016)Wu, Schuster et~al.}]{45610}
Yonghui Wu, Mike Schuster, et~al. 2016.
\newblock {Google's Neural Machine Translation System: Bridging the Gap between
  Human and Machine Translation}.
\newblock \emph{arXiv preprint}, arXiv:1609.08144.

\end{thebibliography}
\bibliographystyle{acl_natbib}

\section*{Appendix}

\subsection*{Parallel document-level corpora}
\label{appendix:websites}

Last access: 19/12/2020

\medskip

\noindent{{WikiHow}:}
\begin{itemize}
    \item \url{https://github.com/esdurmus/Wikilingua}
\end{itemize}

\noindent{{TED-talks}:}
\begin{itemize}
    \item \url{https://object.pouta.csc.fi/OPUS-TED2020/v1/raw/en.zip}
    \item \url{https://object.pouta.csc.fi/OPUS-TED2020/v1/raw/vi.zip}
\end{itemize}

\noindent{{MediaWiki}:}
\begin{itemize}
    \item \url{https://dumps.wikimedia.org/other/contenttranslation} (version: November 2020)

\end{itemize}

\noindent{{OpenSub}:}
\begin{itemize}
    \item \url{https://object.pouta.csc.fi/OPUS-OpenSubtitles/v2018/raw/en.zip}
    \item \url{https://object.pouta.csc.fi/OPUS-OpenSubtitles/v2018/raw/vi.zip}
\end{itemize}

\noindent{{News}:}
\begin{itemize}
    \item \url{https://songngu.dayhoctienganh.net/page/1}
    \item \url{https://toomva.com/doc-bao-anh-viet/pc=8?page=0}
    \item \url{http://cep.com.vn/news}
    \item \url{https://www.hrw.org/languages?language=vi&page=0}
    \item \url{http://vietanhsongngu.com/hoc-tieng-anh-bai-mau-tin-} \url{tuc-c5.htm}
    \item \url{https://baosongngu.com/chuyen-muc/world}
    \\ \url{https://baosongngu.com/chuyen-muc/vn}
    \\ \url{https://baosongngu.com/chuyen-muc/bbc}
    \item \url{https://www.jw.org/vi/tin-tuc/phap-ly}
    \\ \url{https://www.jw.org/vi/tin-tuc/ve-nhan-chung}
    \\ \url{https://www.jw.org/vi/tin-tuc/tin-tuc}
\end{itemize}

\noindent{{Blogspot}:}
\begin{itemize}
    \item \url{https://gocsan.blogspot.com}
\end{itemize}

%% For camera ready...
%\begin{table}[!t]
%\centering
%\resizebox{7.5cm}{!}{
%\setlength{\tabcolsep}{0.33em}
%\begin{tabular}{l|l|l|l|l}
%\hline
%\multirow{2}{*}{\textbf{Domain}} & \multicolumn{2}{c|}{\textbf{Validation}} & \multicolumn{2}{c}{\textbf{Test}} \\
%\cline{2-5}
%& {\textbf{MisA}} & {\textbf{LowQ}} & {\textbf{MisA}} & {\textbf{LowQ}}\\
%\hline
%News & 1 \\ 
%Blogspot & 0 \\
%TED-Talks & 0 \\
%MediaWiki & 7\\ 
%WikiHow & 0 \\
%OpenSubtitles & 24 \\
%\hline
%Total & 32\\
%\hline
%\end{tabular}
%}
%\caption{Misalignment (\textbf{MisA}) and low-quality translation (\textbf{LowQ}) statistics.}
%\label{tab:error}
%\end{table}

%\susection{Notices}

%\subsection*{Discussion on other parallel resources}

%We manually check 100 randomly sampled pairs from the 600K Vietnamese-English sentence pair corpus JW300 \citep{jw300} and find that there are 71 pairs of high-quality translation. However, it is worth noting that JW300 can introduce potential bias because of its religious domain. GNOME from OPUS contains 600K sentences pairs, in which most Vietnamese target sentences include many original translatable technical English words, thus not natural. QED \citep{dataset_qed} has 340K sentence pairs, however, our investigation finds that about a half of the QED pairs are from the TED-Talks transcripts \cite{dataset_ted}; and from the remaining sentence pairs, we randomly sample 100 pairs and find that only 43 pairs have a high-quality translation. 

\subsection*{Discussion on the alignment coverage rate}

OPUS \cite{tiedemann-2012-parallel} also provides a corpus of 350K sentence pairs extracted from a newer version of the MediaWiki content translation dump (version: 02 April 2021),\footnote{\url{https://opus.nlpl.eu/wikimedia-v20210402.php}} wherein duplicate pair removal is not performed. Note that the number of our MediaWiki sentence pairs is bigger at 500K (without taking duplicate pairs into account), thus again showing that our alignment approach is more effective than the OPUS alignment approach.

\subsection*{Discussion on the use of Google Translate}

To align parallel sentences within a parallel document pair as described in Section 2.3, we first translate each English source sentence into Vietnamese by using Google Translate.  Here, the use of Google Translate in this step is via utilizing the ``GoogleTranslate'' function in Google Sheets. However, we later find that this ``GoogleTranslate'' function in Google Sheets produces lower performance scores than using the Google Translate API in both automatic and human evaluation setups. Therefore, in our result tables, we report ``Google Translate'' scores accounted for the Google Translate API on the validation and test sets.

%We also have an extra post-processing sub-step before training/validation/test data split. That is, for  $m$-$n$ sentence alignment where $m \geq 2$ or $n \geq 2$ (e.g. one English source sentence is aligned with two Vietnamese target sentences), we employ two named entity recognition systems  Stanza \cite{qi2020stanza} and PhoNLP \cite{phonlp} for English and Vietnamese, respectively, to determine whether the first token of each sentence is the beginning token of a named entity. 
%\todo{Cite PhoNLP when post-processing, add to appendix}.

\subsection*{Implementation details}
\label{appendix:hparams}

We employ Transformer and mBART implementations from \texttt{fairseq} \citep{fairseq}. % and the official BERT-fused implementation of \citet{bert-nmt} from \texttt{bert-nmt}.\footnote{\url{https://github.com/bert-nmt/bert-nmt}} 
For both Transformer models \cite{transformer}, following \citet{subword_size}, we use \texttt{subword-nmt} to learn joint BPE  with 32K merge operations \citep{subword_bpe}. %\footnote{\url{https://github.com/rsennrich/subword-nmt}}  
For mBART, we fine-tune the pre-trained sequence-to-sequence model mBART25 \cite{mbart}. Here, mBART25 is pre-trained on a Common Crawl dataset of 25 languages, which contains 300GB of English texts and 137 GB of Vietnamese texts. 
Following \citet{transformer}, we use beam search with a beam size of 4 and length normalization of 0.6 for decoding. 
Due to the model size, we apply batch sizes of 16K tokens for Transformer-base, 8K tokens for Transformer-big and 4K tokens for mBART. We optimize the models using Adam \cite{KingmaB14} and run for 30 training epochs, wherein the Adam initial learning rate is warmed up for the first epoch. In addition, we also perform grid search to select the initial learning rate from \{1e-4, 3e-4, 5e-4, 7e-4\} for Transformer models and from \{1e-5, 3e-5, 5e-5, 7e-5\} for mBART.  For both English-to-Vietnamese and Vietnamese-to-English translation setups, the optimal learning rates selected for Transformer-base, Transformer-big and mBART are 5e-4, 3e-4 and 5e-5, respectively. Here, we evaluate each model 8 times during every training epoch, and then select the model checkpoint that obtains the highest BLEU score on the validation set to apply to the test set.

We compute the detokenized and case-sensitive BLEU score using SacreBLEU (with the signature ``BLEU+case.mixed+numrefs.1+smooth.exp+tok.1- 3a+version.1.5.1'').\footnote{\url{https://github.com/mjpost/sacrebleu}} 
Similarly, we also compute the detokenized and case-sensitive TER score (with the option ``-N'' of normalization).\footnote{\url{https://github.com/jhclark/tercom}}

\subsection*{Additional results}

Tables \ref{tab:ter} and  \ref{tab:bleu} present details of TER and BLEU scores on the validation and test sets  for each domain. In addition, we show TER and BLEU scores for each sentence length bucket in Table \ref{tab:ter_sent_len} and Table \ref{tab:bleu_sent_len}, respectively. %We find that models produce poorer scores for short- and medium-length sentences (i.e. $<$ 20 tokens) than for long sentences. This is not surprising as a major proportion of short- and medium-length sentences are from OpenSubtitles, while longer sentences generally come from MediaWiki, News and TED-Talks.

\begin{table*}[!t]
\centering
\resizebox{16cm}{!}{
\setlength{\tabcolsep}{0.33em}
\begin{tabular}{ll|c|c|c|c|c|c|c|c|c|c|c|c}
\hline
& \multirow{2}{*}{\textbf{Model}} & \multicolumn{6}{c|}{\textbf{En-to-Vi}} & \multicolumn{6}{c}{\textbf{Vi-to-En}} \\
\cline{3-14}
& & {\textbf{News}} & {\textbf{BloS}} & {\textbf{TedT}} & {\textbf{MedW}} & {\textbf{WikH}} & {\textbf{OpeS}}  & {\textbf{News}} & {\textbf{BloS}} & {\textbf{TedT}} & {\textbf{MedW}} & {\textbf{WikH}} & {\textbf{OpeS}}  \\
\hline
\multirow{5}{*}{\rotatebox[origin=c]{90}{\textbf{Validation}}} & Google Translate & 35.93 & 47.26 & 49.37 & 24.89 & 51.65 & 60.70 & 42.21 & 51.25 & 46.94 & 30.99 & 51.08 & 50.29 \\
& Bing Translator & 36.07 & 47.47 & 50.51 & 23.17 & 55.53 & 57.43 & 42.72 & 52.38 & 49.46 & 30.17 & 55.92 & 48.57 \\
& Transformer-base & 32.85 & 47.62 & 49.34 & 23.55 & 49.19 & 53.29 & 39.18 & 51.91 & 48.25 & 29.64 & 51.45 & 46.67 \\
& Transformer-big & 32.56 & 46.91 & 48.78 & 22.57 & 48.96 & 52.65 & 37.68 & 50.89 & 47.49 & 29.94 & 50.95 & 46.28 \\
& mBART & \textbf{30.97} & \textbf{45.78} & \textbf{47.87} & \textbf{22.26} & \textbf{47.71} & \textbf{52.67} & \textbf{36.16} & \textbf{48.98} & \textbf{45.75} & \textbf{27.54} & \textbf{49.01} & \textbf{45.56} \\
\hline
\multirow{5}{*}{\rotatebox[origin=c]{90}{\textbf{Test}}} & Google Translate & 45.46 & 53.05 & 46.45 & 27.02 & 53.32 & 60.71 & 49.78 & 56.60 & 45.22 & 33.38 & 52.80 & 51.27 \\
& Bing Translator & 46.44 & 53.96 & 46.83 & 26.03 & 56.28 & 57.62 & 48.61 & 57.89 & 46.95 & 32.48 & 57.15 & 49.28 \\
& Transformer-base & 44.42 & 53.21 & 45.95 & 26.56 & 51.30 & 53.21 & 48.69 & 55.64 & 45.84 & 32.17 & 52.85 & 47.20 \\
& Transformer-big & 43.27 & 52.79 & 45.12 & 25.74 & 50.49 & 52.57 & 47.13 & 54.83 & 45.25 & 32.05 & 52.82 & 47.14 \\
& mBART & \textbf{42.86} & \textbf{51.53} & \textbf{44.84} & \textbf{25.02} & \textbf{50.01} & \textbf{52.44} & \textbf{45.86} & \textbf{54.10} & \textbf{43.57} & \textbf{29.90} & \textbf{51.21} & \textbf{46.16} \\
\hline
\end{tabular}
}
\caption{TER results on each domain.}
\label{tab:ter}
\end{table*}

\begin{table*}[!t]
\centering
\resizebox{16cm}{!}{
\setlength{\tabcolsep}{0.33em}
\begin{tabular}{ll|c|c|c|c|c|c|c|c|c|c|c|c}
\hline
& \multirow{2}{*}{\textbf{Model}} & \multicolumn{6}{c|}{\textbf{En-to-Vi}} & \multicolumn{6}{c}{\textbf{Vi-to-En}} \\
\cline{3-14}
& & {\textbf{News}} & {\textbf{BloS}} & {\textbf{TedT}} & {\textbf{MedW}} & {\textbf{WikH}} & {\textbf{OpeS}}  & {\textbf{News}} & {\textbf{BloS}} & {\textbf{TedT}} & {\textbf{MedW}} & {\textbf{WikH}} & {\textbf{OpeS}}  \\
\hline
\multirow{5}{*}{\rotatebox[origin=c]{90}{\textbf{Validation}}} & Google Translate & 50.28 & 35.67 & 33.62 & 63.43 & 33.60 & 22.37 & 41.12 & 31.85 & 32.21 & 51.32 & 30.48 & 29.18\\
& Bing Translator & 50.13 & 35.66 & 32.64 & 66.10 & 30.19 & 24.89 & 41.02 & 30.95 & 29.51 & 52.90 & 25.88 & 31.30\\
& Transformer-base & 52.63 & 35.24 & 33.51 & 65.60 & 35.32 & 27.42 & 43.54 & 30.61 & 30.67 & 53.50 & 30.39 & 32.87\\
& Transformer-big & 53.39 & 36.17 & 34.00 & 66.96 & 35.80 & 27.95 & 45.35 & 31.84 & 31.83 & 54.11 & 31.11 & 33.60\\
& mBART & \textbf{54.82} & \textbf{36.99} & \textbf{34.72} & \textbf{67.42} & \textbf{37.03} & \textbf{28.18} & \textbf{48.51} & \textbf{33.72} & \textbf{33.76} & \textbf{56.06} & \textbf{33.22} & \textbf{34.93}\\
\hline
\multirow{5}{*}{\rotatebox[origin=c]{90}{\textbf{Test}}} & Google Translate & 41.35 & 31.29 & 36.15 & 61.80 & 30.97 & 23.45 & 34.33 & 26.71 & 34.03 & 48.81 & 28.76 & 28.63\\
& Bing Translator & 40.63 & 30.55 & 35.58 & 63.36 & 28.29 & 25.67 & 35.05 & 25.91 & 32.03 & 50.29 & 24.78 & 30.56\\
& Transformer-base & 41.89 & 29.54 & 36.72 & 62.69 & 32.33 & 28.04 & 34.94 & 27.13 & 33.46 & 50.56 & 28.37 & 32.29\\
& Transformer-big & 43.33 & 30.37 & 37.66 & 63.58 & 33.31 & 28.58 & 36.14 & 28.26 & 34.46 & 51.20 & 28.86 & 32.69\\
& mBART & \textbf{43.93} & \textbf{31.39} & \textbf{38.01} & \textbf{64.67} & \textbf{33.97} & \textbf{29.01} & \textbf{37.04} & \textbf{29.26} & \textbf{36.43} & \textbf{53.93} & \textbf{30.63} & \textbf{34.13}\\
\hline
\end{tabular}
}
\caption{BLEU results on each domain.}
\label{tab:bleu}
\end{table*}

\begin{table*}[!t]
\centering
\resizebox{16cm}{!}{
\setlength{\tabcolsep}{0.33em}
\begin{tabular}{ll|c|c|c|c|c|c|c|c|c|c|c|c}
\hline
& \multirow{3}{*}{\textbf{Model}} & \multicolumn{6}{c|}{\textbf{En-to-Vi}} & \multicolumn{6}{c}{\textbf{Vi-to-En}} \\
\cline{3-14}
& & {\textbf{<10}} & {\textbf{[10, 20)}} & {\textbf{[20, 30)}} & {\textbf{[30, 40)}} & {\textbf{[40, 50)}} & {\textbf{$\geq$ 50}}  & {\textbf{<10}} & {\textbf{[10, 20)}} & {\textbf{[20, 30)}} & {\textbf{[30, 40)}} & {\textbf{[40, 50)}} & {\textbf{$\geq$ 50}}  \\
% \cline{3-14}
 & & 48.77\% & 30.87\% & 13.65\% & 4.46\% & 1.45\% & 0.80\% & 48.77\% & 30.87\% & 13.65\% & 4.46\% & 1.45\% & 0.80\% \\
\hline
\multirow{5}{*}{\rotatebox[origin=c]{90}{\textbf{Validation}}} & Google Translate & 58.35 & 51.61 & 43.33 & 39.31 & 37.77 & 35.68 & 47.78 & 45.44 & 42.72 & 42.68 & 42.62 & 39.93 \\
 & Bing Translator & 54.55 & 51.47 & 44.04 & 39.40 & 38.15 & 35.14 & 46.28 & 47.02 & 44.89 & 42.81 & 42.16 & 38.83 \\
 & Transformer-base & 50.00 & 47.26 & 41.34 & 38.01 & 37.61 & 35.93 & 44.03 & 44.56 & 42.92 & 41.69 & 42.09 & 39.74 \\
 & Transformer-big & 49.64 & 46.68 & 40.64 & 37.46 & 36.79 & 34.88 & 43.60 & 44.04 & 42.64 & 41.40 & 43.20 & 39.29 \\
 & mBART & \textbf{49.77} & \textbf{46.07} & \textbf{39.84} & \textbf{36.74} & \textbf{35.93} & \textbf{34.38} & \textbf{43.16} & \textbf{42.72} & \textbf{40.55} & \textbf{38.76} & \textbf{39.11} & \textbf{36.18} \\
\hline
\multirow{5}{*}{\rotatebox[origin=c]{90}{\textbf{Test}}} & Google Translate & 59.34 & 52.68 & 44.78 & 40.29 & 39.37 & 37.83 & 48.80 & 47.05 & 43.87 & 43.84 & 43.56 & 43.39 \\
 & Bing Translator & 55.24 & 52.28 & 44.86 & 40.50 & 40.18 & 38.22 & 46.86 & 48.57 & 45.37 & 43.58 & 41.84 & 41.76 \\
 & Transformer-base & 50.40 & 48.50 & 42.48 & 39.67 & 39.74 & 38.59 & 44.13 & 46.15 & 43.61 & 42.98 & 42.45 & 41.98 \\
 & Transformer-big & 49.92 & 47.84 & 41.89 & 38.93 & 38.81 & 37.26 & 44.00 & 45.77 & 43.34 & 42.25 & 43.78 & 42.12 \\
 & mBART & \textbf{50.01} & \textbf{47.52} & \textbf{41.33} & \textbf{37.98} & \textbf{38.43} & \textbf{36.60} & \textbf{43.32} & \textbf{44.52} & \textbf{41.73} & \textbf{40.40} & \textbf{39.71} & \textbf{39.35} \\
\hline
\end{tabular}
}
\caption{TER results w.r.t. sentence lengths of reference English sentences.}
\label{tab:ter_sent_len}
\end{table*}

\begin{table*}[!t]
\centering
\resizebox{16cm}{!}{
\setlength{\tabcolsep}{0.33em}
\begin{tabular}{ll|c|c|c|c|c|c|c|c|c|c|c|c}
\hline
& \multirow{3}{*}{\textbf{Model}} & \multicolumn{6}{c|}{\textbf{En-to-Vi}} & \multicolumn{6}{c}{\textbf{Vi-to-En}} \\
\cline{3-14}
& & {\textbf{<10}} & {\textbf{[10, 20)}} & {\textbf{[20, 30)}} & {\textbf{[30, 40)}} & {\textbf{[40, 50)}} & {\textbf{$\geq$ 50}}  & {\textbf{<10}} & {\textbf{[10, 20)}} & {\textbf{[20, 30)}} & {\textbf{[30, 40)}} & {\textbf{[40, 50)}} & {\textbf{$\geq$ 50}}  \\
% \cline{3-14}
 & & 48.77\% & 30.87\% & 13.65\% & 4.46\% & 1.45\% & 0.80\% & 48.77\% & 30.87\% & 13.65\% & 4.46\% & 1.45\% & 0.80\% \\
\hline
\multirow{5}{*}{\rotatebox[origin=c]{90}{\textbf{Validation}}} & Google Translate & 26.64 & 38.72 & 44.62 & 48.28 & 49.72 & 52.05 & 32.06 & 35.95 & 39.01 & 39.43 & 39.70 & 41.75 \\
 & Bing Translator & 29.23 & 38.93 & 44.76 & 48.37 & 49.04 & 53.15 & 34.26 & 34.83 & 37.33 & 39.43 & 40.31 & 43.68 \\
 & Transformer-base & 31.48 & 41.65 & 47.22 & 49.44 & 49.19 & 52.01 & 36.28 & 36.87 & 39.32 & 40.41 & 39.89 & 43.16 \\
 & Transformer-big & 31.92 & 42.29 & 47.90 & 49.98 & 50.67 & 53.59 & 36.94 & 37.69 & 40.03 & 41.04 & 41.82 & 43.75 \\
 & mBART & \textbf{32.73} & \textbf{42.95} & \textbf{48.63} & \textbf{50.65} & \textbf{51.40} & \textbf{ 54.08} & \textbf{37.99} & \textbf{39.60} & \textbf{41.90} & \textbf{43.64} & \textbf{43.31} & \textbf{46.29} \\
\hline
\multirow{5}{*}{\rotatebox[origin=c]{90}{\textbf{Test}}} & Google Translate & 27.00 & 37.17 & 44.20 & 46.58 & 48.52 & 50.57 & 32.27 & 34.24 & 37.47 & 38.17 & 37.94 & 38.30 \\
 & Bing Translator & 29.68 & 37.46 & 44.07 & 46.23 & 48.12 & 50.17 & 34.20 & 32.97 & 36.66 & 38.27 & 39.73 & 40.70 \\
 & Transformer-base & 32.03 & 39.40 & 45.86 & 47.16 & 48.18 & 48.76 & 36.78 & 34.99 & 38.27 & 38.86 & 38.76 & 39.46 \\
 & Transformer-big & 32.36 & 40.48 & 46.45 & 48.13 & 49.39 & 50.18 & 37.23 & 35.49 & 39.02 & 39.90 & 39.34 & 39.85 \\
 & mBART & \textbf{33.15} & \textbf{40.93} & \textbf{47.35} & \textbf{48.68} & \textbf{49.80} & \textbf{51.15} & \textbf{38.31} & \textbf{37.22} & \textbf{41.05} & \textbf{42.21} & \textbf{42.49} & \textbf{43.35} \\
\hline
\end{tabular}
}
\caption{BLEU results w.r.t. sentence lengths of reference English sentences.}
\label{tab:bleu_sent_len}
\end{table*}

\end{document}